\pdfoutput=1
\documentclass{article} 
\usepackage{iclr2027_conference,times}
\usepackage{graphicx}

\usepackage{amsmath,amsfonts,bm}

\def\eqref#1{equation~\ref{#1}}

\def\1{\bm{1}}

\DeclareMathAlphabet{\mathsfit}{\encodingdefault}{\sfdefault}{m}{sl}
\SetMathAlphabet{\mathsfit}{bold}{\encodingdefault}{\sfdefault}{bx}{n}

\usepackage{hyperref}
\usepackage{url}
\usepackage{booktabs}
\usepackage{amsthm}
\usepackage{amssymb}

\newtheorem{proposition}{Proposition}
\newtheorem{corollary}{Corollary}
\theoremstyle{definition}
\newtheorem{definition}{Definition}

\title{What Does It Mean to Remove an Edge? \\Intervention-Aware Fidelity for Graph Neural Network Explanations}

\author{Rocco Mattia Di Mauro\thanks{Equal contribution.}, Marco Grassia\footnotemark[1] \& Giuseppe Mangioni \\
Department of Electrical, Electronic and Computer Engineering (DIEEI) \\
University of Catania, Italy \\
\texttt{rocco.dimauro@phd.unict.it}, \\ \texttt{\{marco.grassia,giuseppe.mangioni\}@unict.it}
}

\iclrfinalcopy

\begin{document}

\maketitle
    
\begin{abstract}
Graph neural network explanations are often evaluated via deletion fidelity, i.e., by deleting their top-ranked edges and measuring the resulting change in model output.
Yet, in a message-passing GNN, “deleting an edge” does not specify a unique intervention. 
Structural deletion removes the edge from the graph and recomputes quantities such as degree normalization and attention, whereas message suppression leaves this structural context intact while blocking the message carried by the edge. 
The meaning of deletion fidelity therefore depends on what the intervention targets in the model computation.
We formalize this distinction and derive sufficient conditions for exact equivalence. 
The two interventions coincide for additive message passing with edge-set-independent coefficients, but can diverge when aggregation depends on the presence of zero messages or when removing an edge changes the coefficients applied to surviving messages.
We then construct intervention-specific edge attributions and evaluate each resulting ranking under both interventions, testing whether their relative fidelity changes with the evaluation intervention. 
Across nine datasets and a panel of GIN, GCN, and GATv2 models at depths 2–5, all nine GIN controls satisfy the equivalence predicted by the theory.
On the target logit, the relative fidelity of the two rankings depends on the evaluation intervention in 10 of 11 reportable GCN and GATv2 cells, and in nine of them each ranking has the larger deletion area under the intervention it was built with.
Deletion fidelity is thus not a property of an edge ranking alone. 
Evaluation protocols should explicitly report the intervention, output functional, target, and perturbation unit, and test alternative intervention semantics when the scientific question does not uniquely determine one.
\end{abstract}

\section{Introduction}
\begin{figure}[t]
\centering
\includegraphics[width=\textwidth]{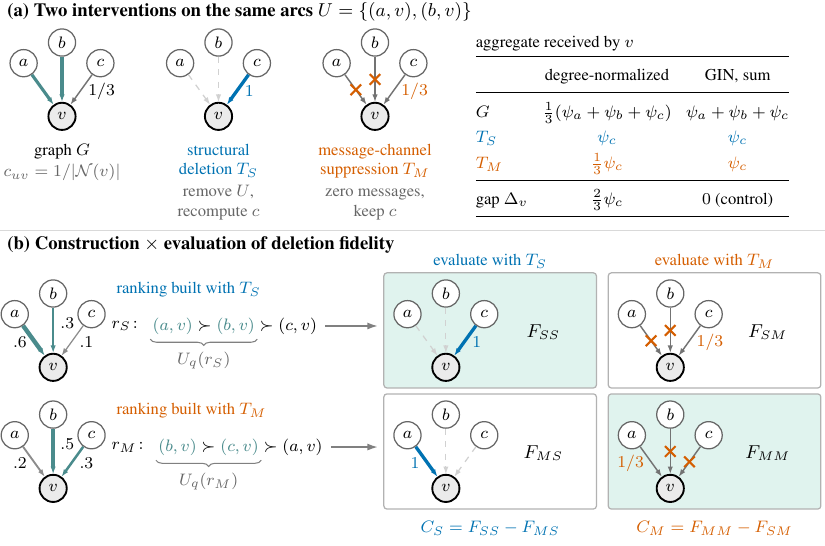}
\caption{\textbf{Removing an edge can mean two different interventions, and both the explainer and the evaluator choose one.}
\textbf{(a)}~Structural deletion $T_S$ removes the arcs $U$ into $v$ (teal) from the graph and recomputes the coefficients of the arcs that survive; message-channel suppression $T_M$ sets the messages on $U$ to zero and keeps every coefficient.
In a degree-normalized layer, $c_{uv}=1/|\mathcal N(v)|$ with the self-loop omitted, so the surviving arc $(c,v)$ has coefficient $1$ under $T_S$ and $1/3$ under $T_M$.
The table gives the aggregate received by $v$, where $\psi_u$ is the message from $u$.
The gap between the $T_S$ and $T_M$ rows is $\Delta_v$ of Equation~\ref{eq:divergence}.
It vanishes under sum aggregation with unit coefficients (GIN, Proposition~\ref{prop:equivalence}), which makes GIN a control.
\textbf{(b)}~An explainer built with $T_S$ or $T_M$ scores the arcs (illustrative values), and sorting the scores gives the ranking $r_S$ or $r_M$.
Each ranking is evaluated under both interventions by removing its top-$q$ prefix $U_q(r)$.
$F_{\mathcal{O}\mathcal{P}}$ is the deletion fidelity of the ranking built with $T_\mathcal{O}$ and evaluated with $T_\mathcal{P}$, i.e., the area under the output-change curve as $q$ grows.
Shaded boxes are the matched pairs, in which construction and evaluation use the same intervention.
Each contrast compares the two rankings within one evaluation column, and their sum is the interaction $D=C_S+C_M$.}
\label{fig:overview}
\end{figure}

Graph Neural Networks (GNNs) have emerged as the dominant paradigm for learning from relational data.
As their adoption continues to grow, explaining their predictions remains a significant challenge.
The most common answer is an edge-level attribution: a score per edge that ranks how much each connection contributes to the model output, from which an explanation subgraph can be extracted.
Yet, an attribution is meaningful only relative to a question about the model.
Fidelity metrics pose such questions by removing edges.
Sufficiency asks whether the top-ranked edges preserve the model prediction when all the others are removed, whereas deletion fidelity asks how the prediction changes as top-ranked edges are removed.
Both require no ground-truth explanation, but the word ``removed'' leaves the intervention unspecified.

Structural deletion changes the edge set and therefore every quantity computed from it, from node degrees and neighborhoods to degree normalization in GCN~\citep{kipf2017semisupervised} or attention in GAT layers~\citep{velickovic2018graph}.
By contrast, message-channel suppression keeps the edges but zeroes the selected transmitted messages.
Both interventions suppress the selected direct messages, but they can change the contributions of surviving edges differently.
This ambiguity enters twice: an explainer may use one intervention to construct a ranking, while the evaluator may use either intervention to score it (Figure~\ref{fig:overview}).

We ask whether the intervention an explainer uses to \emph{build} its ranking interacts with the intervention used to \emph{evaluate} it, and we make four contributions.
First, we give sufficient conditions for the two interventions to coincide, namely an aggregator that absorbs zero messages and coefficients that do not depend on the edge set (Section~\ref{sec:preliminaries}).
GIN satisfies both, GCN and GATv2 violate the second, and mean aggregation violates the first.
With a sum aggregator, the first-layer gap is carried entirely by the renormalized surviving messages.
Second, we cross the two interventions in a construction-by-evaluation design, whose two matched contrasts and interaction are estimated with training seeds as clusters and with GIN as a built-in control (Section~\ref{sec:framework}).
Third, we apply the design to edge Integrated Gradients on 22 dataset--backbone combinations, which we call \emph{cells} (Section~\ref{sec:results}).
All nine GIN controls agree to within $10^{-5}$ on the target logit, as the theory predicts.
On the same functional, the interaction is positive in 10 of the 11 reportable GCN and GATv2 cells, and in nine of them each ranking has the larger deletion area under the intervention it was built with.
The verdict can change with the output functional and with the perturbation unit.
Fourth, we repeat the crossing with GNNExplainer on the same 11 cells.
A fixed learned ranking scores differently under the two interventions in seven cells on the target logit, but in only four does each learned mask have the larger deletion area under its own intervention, and restarts of the explainer often disagree, so this analysis is exploratory.
We do not claim that either intervention is the correct one, that a larger deletion area means a better explanation, or that our results rank explainers.
We recommend reporting fidelity together with its intervention, output functional, target and perturbation unit, and crossing the two interventions whenever the scientific question does not fix one.

\section{Background and Related Work}
\paragraph{Fidelity and its critiques.}
Without ground truth, explanations are scored by perturbing the input and reading the change in model output~\citep{sanchez2020evaluating,amara2022graphframex,agarwal2023graphxai}.
Necessity ($\mathrm{Fid}+$), or deletion fidelity, removes the explanation and measures the output drop; sufficiency ($\mathrm{Fid}-$) keeps only the explanation.
Plausibility is scored separately, by recovery of annotated ground-truth structures.
Several works have shown that these scores depend on how the perturbation is implemented.
Because removal produces graphs outside the training distribution, GInX-Eval retrains the model on perturbed graphs~\citep{amara2023ginxeval}, while Robust Fidelity and F-Fidelity randomize or fine-tune the evaluation~\citep{zheng2024robustfidelity,zheng2025ffidelity}.
\citet{azzolin2025faithfulness} show that faithfulness metrics are not interchangeable, and \citet{azzolin2026donotexplain} construct explanations that score well without explaining the model.
At the level of the computation, GraphMask~\citep{schlichtkrull2021interpreting} masks messages rather than edges, and FlowX~\citep{flowx} scores layer-wise message flows.
All these works act on the \emph{evaluation} perturbation and keep the explanation fixed.
We look at the other side of the pipeline, since the explainer also perturbs the model while it \emph{constructs} the explanation, and we cross the two choices.

\paragraph{GNN explainers.}
Integrated Gradients (IG)~\citep{sundararajan2017axiomatic} integrates the gradient of the output along a path from a baseline to the input.
Applied to a continuous gate on each edge, it yields edge scores that sum to the output change between the fully gated and the intact graph.
GNNExplainer~\citep{gnnexplainer} and PGExplainer~\citep{luo2020parameterized} learn a soft edge mask that preserves the prediction under a size penalty.
SubgraphX~\citep{subgraphX} searches connected, node-induced subgraphs; its sparsity is defined over selected nodes, so it does not yield an edge ranking comparable to an edge mask without an additional conversion.
IG and mask-based explainers perturb the model while constructing their explanation, so the explainer chooses an intervention before the evaluator chooses one again.

\section{Preliminaries}
\label{sec:preliminaries}
Let $G=(V,E)$ be a graph with $N=|V|$ nodes and a set $E \subseteq V\times V$ of directed edges, which we call \emph{arcs}, with one arc in each direction for an undirected edge, and let $X\in\mathbb{R}^{N\times F}$ be the node feature matrix, where $F$ is the feature dimension.
For a multi-class classification task, a GNN-based classifier is a map $f:(G,X)\to\mathbb{R}^{O\times C}$ that assigns each of its $O$ output instances one logit for each of the $C$ classes; $O$ is $N$, $|E|$, or $1$ for node, edge, and graph classification, respectively.
We focus on explaining the model output associated with the original model's predicted class $c$ for a single instance in a supervised learning setting.
Accordingly, the function $f_c:(G,X)\to\mathbb{R}^O$ becomes the primary object of analysis.
For our graph-classification experiments, $f_c(G,X)$ denotes the target-class logit of the selected graph, with $c$ fixed to the intact model's predicted class before any perturbation. Let $U\subseteq E$ be the perturbed arcs and $r$ an attribution ranking. At removal fraction $q\in[0,1]$, $U_q(r)$ consists of the top-ranked $\lceil q|E|\rceil$ arcs, with zero selected at $q=0$ and exact ties resolved by ascending arc index in the primary analysis.
Let $T_S(G,U)$ denote the transformation that removes the edges $e\in U$ from the graph $G$.
Following this operation, all graph-dependent quantities are recomputed, including node degrees, neighborhood structures, normalization factors, attention coefficients, and any other mechanism whose computation depends on the graph connectivity or the number of edges.
Finally, let $T_M(G,U)$ denote the transformation that preserves the topology of the graph while silencing the messages transmitted along the edges $e\in U$.
\paragraph{When the two interventions coincide.}
To compare the two interventions we write a message-passing layer as
\begin{equation}
\label{eq:layer}
h^{(l+1)}_v=\phi^{(l)}\Big(h^{(l)}_v,\;\mathfrak{A}_{(u,v)\in E}\big[c^{(l)}_{uv}(E,H^{(l)})\,\psi^{(l)}(h^{(l)}_u,h^{(l)}_v)\big]\Big),
\end{equation}
Here $h^{(l)}_v$ is node $v$'s row in the hidden-state matrix $H^{(l)}$, with $x_v=h_v^{(0)}$ in $X$; $\psi^{(l)}$ is the message function, $\mathfrak{A}$ is a permutation-invariant aggregator, $\phi^{(l)}$ updates the node state, and $c^{(l)}_{uv}(E,H^{(l)})$ is a coefficient that may depend on the graph and hidden states. Both the aggregator and the coefficients can make the contribution of one arc depend on other arcs.
GIN uses sum aggregation with $c\equiv1$.
GCN uses $c_{uv}=(\tilde d_u\tilde d_v)^{-1/2}$, with degrees $\tilde d$ counted on $E$ plus self-loops.
GATv2 uses a softmax over the neighborhood of $v$.
$T_S$ recomputes every coefficient on $E\setminus U$, whereas $T_M$ keeps the coefficients computed on $E$ and sets the messages on $U$ to zero.
Two properties of a layer decide whether this difference matters.

\begin{definition}
The aggregator \emph{absorbs zero} if adding zero messages never changes its value.
The coefficients are \emph{edge-set free} if $c^{(l)}_{uv}(E,H^{(l)})=c^{(l)}_{uv}(E\setminus U,H^{(l)})$ for every $U\subseteq E$, every hidden-state matrix $H^{(l)}$ and every surviving arc $(u,v)\in E\setminus U$.
\end{definition}

A sum absorbs zero and a mean does not, since the zeros enter its denominator.
Unit coefficients are edge-set free, while degree normalization and attention are not, since removing an edge changes the degree or the softmax denominator seen by the surviving ones.

\begin{proposition}
\label{prop:equivalence}
If every layer of the form in Equation~\ref{eq:layer} absorbs zero and has edge-set-free coefficients, and the readout depends only on the final node states, then $f(T_S(G,U))=f(T_M(G,U))$ for every graph, every parameter setting and every $U\subseteq E$.
\end{proposition}

The proof is an induction over layers (Appendix~\ref{app:theory}).
The conditions are sufficient but not necessary, because the two failures can compensate each other in contrived layers.

\begin{corollary}
\label{cor:gin}
For GIN with sum aggregation, $T_S$ and $T_M$ coincide exactly, so every GIN model is a control on which any construction-by-evaluation effect must vanish.
\end{corollary}

GCN and GATv2 violate edge-set freeness, and mean aggregation violates zero absorption.
In both cases the interventions can diverge, and Appendix~\ref{app:theory} gives explicit witnesses for each failure mode.
When the aggregator is a sum, the divergence at the first layer has a closed form.

\begin{proposition}
\label{prop:divergence}
With $\mathfrak{A}=\sum$, the first-layer aggregates of node $v$ under $T_S$ and $T_M$ differ by
\begin{equation}
\label{eq:divergence}
\Delta_v=\sum_{(u,v)\in E\setminus U}\big[c^{(0)}_{uv}(E\setminus U,X)-c^{(0)}_{uv}(E,X)\big]\,\psi^{(0)}(x_u,x_v).
\end{equation}
\end{proposition}
The set $E\setminus U$ in Equation~\ref{eq:divergence} consists of the arcs that survive structural deletion. Arcs in $U$ contribute nothing to either first-layer sum: they are absent under $T_S$ and their messages are zero under $T_M$.
The gap is carried entirely by the \emph{surviving} messages, whose coefficients are renormalized under $T_S$ and left unchanged under $T_M$.
Figure~\ref{fig:overview} shows this on a single node.
The bound $\|\Delta_v\|_\infty\le R_v(U)\max\|\psi_{uv}\|_\infty$, with $R_v(U)$ the total coefficient shift, gives a zero condition but not a dose--response, since shifted messages can cancel (Appendix~\ref{app:theory}).

\section{Intervention-Specific Attribution}
\label{sec:attribution}
IG attributes the change from a baseline to the input by integrating the gradient along a continuous path.
A continuous gate $g_e\in[0,1]$ is assigned to each directed arc. The structural construction injects that gate before GCN degree normalization or before the GATv2 attention softmax; the message-channel construction multiplies the transmitted message after the corresponding context is computed. Write $f_c^{\mathcal O}(g)$ for the target logit along the gate path of construction $\mathcal O\in\{S,M\}$. The two paths can differ even when they start at the same intact model, so their integrated gradients can give different rankings.
For each construction, signed IG has a path-specific completeness identity, up to numerical quadrature error:
\begin{equation}
    \sum_e IG^{\mathcal O}_e \approx f_c^\mathcal O(\mathbf 1)-f_c^\mathcal O(\mathbf 0).
\end{equation}

The same choice exists inside GNNExplainer.
Its soft edge mask can multiply the message after normalization, as in the PyTorch Geometric implementation~\citep{fey2019pyg}, or enter the edge weights before degrees and attention are computed.
The first construction targets the message channel and the second the structure.
Since the objective is optimized through the mask, the two constructions generally return different rankings.
We therefore run GNNExplainer in both forms as an exploratory second instance of the design, and also evaluate the PyG ranking alone under both interventions.

\section{The Evaluation Framework}
\label{sec:framework}

Let $\mathcal O\in\{S,M\}$ identify the edge-IG construction and $\mathcal P\in\{S,M\}$ the deletion evaluator. The two choices form a $2\times2$ design.
For a fixed target class $c$, let $\boldsymbol z(G,X)$ be the intact logit vector and let $\boldsymbol z(T_{\mathcal P}(G,U),X)$ be the logits after evaluation intervention $\mathcal P$ on arcs $U$. We read three scalar output functionals from these logits: target logit $\Phi_c^{\mathrm{logit}}(\boldsymbol z)=z_c$, target-versus-rest margin $\Phi_c^{\mathrm{margin}}(\boldsymbol z)=z_c-\log\sum_{k\ne c}e^{z_k}$, and target probability $\Phi_c^{\mathrm{prob}}(\boldsymbol z)=e^{z_c}/\sum_k e^{z_k}$. The target $c$ is never recomputed after intervention.
Let $r_{\mathcal O}$ be the signed arc ranking built using $\mathcal O$, and let $U_q(r_{\mathcal O})$ be its top-$q$ prefix. Our necessity-style deletion response and its unnormalized area are
\begin{equation}
\label{Eq:fidelity}
    F_{\mathcal O\mathcal P}(q;\Phi_c)
    =\Phi_c\!\big(\boldsymbol z(G,X)\big)
     -\Phi_c\!\big(\boldsymbol z(T_{\mathcal P}(G,U_q(r_{\mathcal O})),X)\big),
    \qquad
    F_{\mathcal O\mathcal P}
    =\int_{0}^{0.5}F_{\mathcal O\mathcal P}(q;\Phi_c)\,dq.
\end{equation}
The integral is evaluated on the grid $q\in\{0,0.1,\ldots,0.5\}$ by the trapezoidal rule. The first index changes the ranking; the second changes the model computation used to score that same prefix.

Using the deletion areas defined in Eq.~\ref{Eq:fidelity}, we introduce two contrasts. 
The first,
\begin{equation*}
    C_S=F_{SS}-F_{MS},
\end{equation*}
compares structural and message-channel rankings under a fixed structural evaluator. 
The second,
\begin{equation*}
    C_M=F_{MM}-F_{SM},
\end{equation*}
compares the same rankings under a fixed message-channel evaluator.
Thus, $C_S$ and $C_M$ isolate the contribution of the construction under each evaluator.

From the two contrasts, we obtain the interaction $D$, a difference-in-differences:
\begin{equation*}
    D=C_S+C_M.
\end{equation*}
Algebraically, $D=(F_{SS}-F_{MS})-(F_{SM}-F_{MM})$: it compares the structural ranking's advantage under the two evaluators. Thus $D>0$ means that this advantage is larger under structural evaluation. It does not require either contrast to be positive; for example, $C_S=2$ and $C_M=-1$ give $D=1$ although the structural ranking has the larger area under both evaluators. A nonzero $D$ is an interaction in measured fidelity, not an interaction in the model output.
Accordingly, we define matched superiority as the simultaneous positivity of the two contrasts:
\begin{equation*}
    C_S > 0 \land C_M > 0.
\end{equation*}
This strongest condition indicates that each ranking has the larger deletion area under the evaluator that matches its construction.

Another important factor concerns how edge attribution scores are translated into the ranking $r$.
Message passing implements an undirected edge $\{u,v\}$ as two arcs, $(u,v)$ and $(v,u)$, and an edge attribution method may score each arc separately.
We call this the \emph{arc} unit: each arc is ranked and removed on its own, which doubles the number of ranked elements in an undirected graph.
The \emph{bond} unit treats the two arcs as one element, so removing a bond affects both message directions.
Its IG score is the sum of the scores of its two arcs, $\mathrm{IG}_{\{u,v\}}=\mathrm{IG}_{(u,v)}+\mathrm{IG}_{(v,u)}$, which preserves completeness: the bond scores still sum to the output change explained by IG.

\section{Experimental Design}
\label{sec:design}
\paragraph{Datasets and models.}
We evaluate the framework on nine graph-classification datasets: four synthetic benchmarks with edge-level ground truth (BA-2Motifs, BA-MultiShapes, SPMotif, and a custom synthetic distance task, Synth-Dist, described in Appendix~\ref{app:protocol}) and five real-world datasets (BACE, MUTAG, Mutagenicity, Reddit-Binary, Tox21). 
All are graph-classification tasks.
The tasks are addressed with three GNN backbones spanning the message-passing spectrum: GIN~\cite{xu2018how} (post-aggregation message-sum), GCN~\cite{kipf2017semisupervised} (degree normalization), and GATv2~\cite{brody2022how} (attention).
Each dataset--backbone cell is trained with $10$ random seeds at each depth $K\in\{2,3,4,5\}$ ($40$ models per cell), and every trained model is evaluated on up to $50$ test graphs.
For each graph, the target is the original model's predicted class, fixed before any intervention.

\paragraph{Attribution and crossed evaluation.}
We use one edge IG method with two gate injection points.
Structural attribution applies the gate before graph-dependent normalization or attention, whereas message-channel attribution applies it to the transmitted message after that context has been computed.
Both constructions use the same number of IG steps for each trained model.
We select this number on validation graphs, checking each construction against its own full-removal endpoint.
We rank directed arcs by descending signed attribution, resolving exact ties by ascending arc index to ensure reproducibility.

Each ranking is evaluated under both structural deletion and message-channel suppression.
At removal fractions $q\in\{0,0.1,\ldots,0.5\}$, we remove the top-ranked fraction of arcs and record the change from the unperturbed model output. 
Fidelity is the unnormalized trapezoidal area under this deletion-response curve (Equation~\ref{Eq:fidelity}).
The target logit is the primary output functional; the
target-versus-rest margin and target probability are secondary
functionals of the same perturbed forward passes.

\paragraph{Estimand and inference.}
For each graph, the four \emph{arms} of the crossed design yield  four deletion areas $F_{SS}$, $F_{SM}$, $F_{MS}$ and $F_{MM}$; their interaction is $D$.
We first average graph-level values within each trained model, then average the four depths equally within each seed.
The cell estimate is the mean of the ten seed-level values.
We obtain 95\% percentile bootstrap intervals by resampling seeds, keeping depths paired within each seed. 
Graphs and removal fractions are not treated as independent training replicates. 
We test $D\neq0$ with an exact two-sided sign-flip test over the ten seed-level values.
The $p$-values are corrected with the Benjamini--Yekutieli procedure~\citep{benjamini2001control} within two families, never across them.
The primary family holds the three cells declared before the full depth-by-seed crossing was analyzed (BA-2Motifs--GCN, BACE--GCN and BACE--GATv2), and the secondary family holds the remaining reportable cells.
Matched superiority, $C_S>0\land C_M>0$, is declared when both one-sided 95\% bootstrap lower bounds are above zero.

\paragraph{Exploratory GNNExplainer evaluator check.}
We use PyG-GNNExplainer's message-channel edge mask with node masking disabled, 100 optimization epochs and five fixed restarts per graph. For each learned ranking, we hold the ranking and its target fixed and compute deletion areas under both evaluators. All 11 reportable GCN and GATv2 cells have complete fixed-budget summaries. We average restart-level paired differences within graph, then graph-level differences within each trained model, and balance depths within training seed. We report 95\% percentile seed-cluster bootstrap intervals without multiplicity adjustment. This check describes the fixed-budget rankings and does not estimate a GNNExplainer construction-by-evaluation interaction (Section~\ref{sec:limitations}).
For the descriptive Figure~\ref{fig:contrasts}b, the same run also optimizes a structural mask that enters before GCN degree normalization or GATv2 attention normalization. Within each graph and restart the two constructions share checkpoint, target, initial mask seed, objective and 100-epoch budget; each ranking is then evaluated under both interventions. The fixed-ranking check above is the PyG arm of this run, with gap $F_{MM}-F_{MS}$. The plotted 90\% seed-cluster intervals are recomputed from depth-balanced seed means, without multiplicity adjustment.

\paragraph{Controls and sensitivity analyses.}
GIN provides the equivalence control: we check agreement between the two evaluators directly, since a zero difference-in-differences alone could arise by cancellation.
We additionally examine the two secondary output functionals, random resolutions of attribution ties, and undirected bonds formed by grouping the two directed arcs of an edge.
Model-health and attribution-completeness checks determine which cells can support numerical fidelity claims; excluded cells remain visible in the reported panel.

\section{Results}

\label{sec:results}

\paragraph{Controlled construction-by-evaluation crossing.}
The edge-IG crossing covers 22 dataset--backbone cells, and the nine GIN cells are equivalence controls.
The three Tox21 classifiers are not informative enough for a fidelity claim, so the interaction analysis comprises eight GCN and three GATv2 cells.

\begin{figure}[t]
\centering
\includegraphics[width=\textwidth]{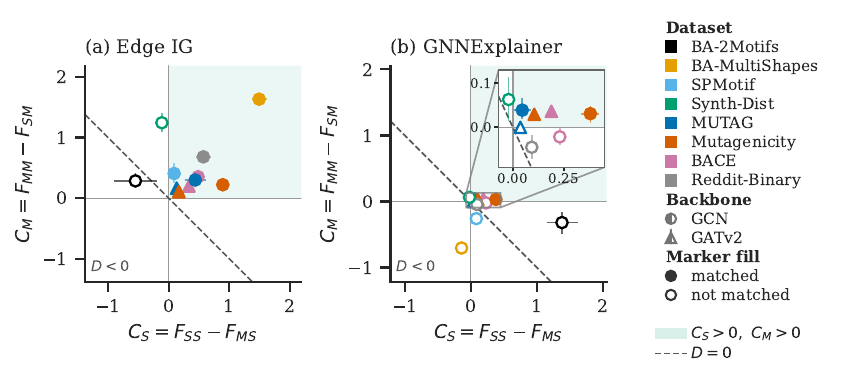}
\caption{\textbf{Matched contrasts on the target logit.}
The horizontal axis is $C_S=F_{SS}-F_{MS}$ and the vertical axis is $C_M=F_{MM}-F_{SM}$; cells above the dashed line have $D=C_S+C_M>0$.
Bars are 90\% percentile seed-cluster bootstrap intervals, and filled markers meet the two one-sided 95\% lower-bound criterion for matched superiority.
\textbf{(a)}~The controlled edge-IG crossing in 11 reportable GCN and GATv2 cells. The GIN equivalence controls lie at the origin and are omitted for clarity.
\textbf{(b)}~An exploratory crossing of fixed-budget PyG message-channel and structural GNNExplainer masks in the same 11 cells. The marker criterion describes these particular 100-epoch, five-restart rankings; the restart-stability pilot did not certify a converged ranking budget, so this panel does not establish a general GNNExplainer construction effect. The inset magnifies the cells near the origin.}
\label{fig:contrasts}
\end{figure}

Figure~\ref{fig:contrasts}a shows the two matched contrasts for the edge-IG crossing; Table~\ref{tab:operator-crossing-results} in Appendix~\ref{app:tables} lists the target-logit interactions.
In the primary family, BACE--GCN and BACE--GATv2 have a positive interaction ($p_{\mathrm{BY}}=0.005$) and BA-2Motifs--GCN does not.
All eight secondary cells have a positive interaction ($p_{\mathrm{BY}}=0.005$), so $D>0$ holds in 10 of the 11 cells.
Matched superiority holds in nine of them.
The distinction matters in Synth-Dist--GCN: its interaction is positive, but $C_S=-0.112$ and $C_M=1.243$, so the message-channel ranking has the larger deletion area even under structural evaluation.
In BA-2Motifs--GCN, the mean interaction is $-0.263$ (95\% CI $[-0.561,0.064]$); depth-specific estimates change from $+0.43$ at $K=2$ to $-0.18$, $-0.92$, and $-0.38$ at $K=3,4,5$.
Thus the pooled result does not describe a depth-invariant effect in this cell.

\paragraph{Equivalence controls and sensitivity to the reported quantity.}
Corollary~\ref{cor:gin} predicts a zero gap between the two evaluators on every GIN cell.
Across the nine GIN controls, the largest absolute gap is $9.28\times10^{-6}$ on the target logit, $1.82\times10^{-5}$ on the margin and $3.70\times10^{-7}$ on the probability, consistent with floating-point error.

The interaction is not invariant to the output functional (Table~\ref{tab:crossing-functionals}).
Among the 11 cells, $D>0$ holds in 10 on the logit, nine on the margin and ten on the probability, and matched superiority in nine, eight and nine.
The cells that change differ by functional.
SPMotif--GCN loses the interaction on the margin and on the probability, while BA-2Motifs--GCN gains it on the probability, with $D=0.0426$ (95\% CI $[0.0287,0.0597]$).
The three functionals are therefore not rescalings of one another.

The perturbation unit also changes the question.
When the two directed arcs of an undirected edge are grouped into a bond and their signed IG scores are summed, the descriptive sensitivity supports a logit interaction and matched superiority in all 11 cells, compared with 10 and nine at the arc unit.
In particular, BA-2Motifs--GCN changes from an unestablished arc interaction to a positive bond interaction of $0.332$ (95\% CI $[0.177,0.488]$). 
Twenty random tie-break priorities plus the deterministic priority changed none of the cell verdicts.

\paragraph{Exploratory GNNExplainer checks.}
Figure~\ref{fig:contrasts}b repeats the crossing with two GNNExplainer masks, one built through PyG's message-channel masking and one through structural deletion. Matched superiority holds in four of the 11 cells (Table~\ref{tab:gnnexplainer-crossing}), against nine for edge IG. Restarts of GNNExplainer on the same graph often produce weakly correlated rankings, so this panel describes the measured rankings rather than a stable property of either construction.
Table~\ref{tab:gnnexplainer-evaluator} instead holds each PyG-GNNExplainer ranking fixed and compares its deletion area under message-channel and structural evaluation. The 95\% seed-cluster interval for the paired gap excludes zero in seven of the 11 cells on the logit, eight on the margin and nine on the probability; the signs vary by cell and functional. For example, BA-2Motifs--GCN has a positive logit gap and a negative probability gap. This paired comparison remains interpretable for the measured restart draws. Appendix~\ref{app:tables} also reports a matched-size structural comparison with SubgraphX.

\section{Limitations}
\label{sec:limitations}

\paragraph{Meaning of the counterfactuals.}
Structural deletion and message-channel suppression answer different questions about a fixed model computation. 
Neither intervention is universally preferable: the appropriate intervention depends on whether the question concerns removal of a graph relation or interruption of its transmitted message while retaining the structural context. 
Structural deletion can create out-of-distribution graphs, including chemically implausible molecular structures, and message-channel suppression can produce internal computations absent from training.
A large deletion area shows sensitivity of the selected output to the specified intervention; by itself it does not establish recovery of an annotated rationale or a causal effect in the data-generating process.

\paragraph{Panel and inference.}
The empirical panel consists of graph-classification benchmarks and frozen trained models, rather than a random sample of graph domains or GNN architectures. 
Only three GATv2 cells enter the interaction analysis, all on molecular datasets, so architecture and domain effects cannot be separated in that stratum. 
Training seed is the inferential cluster; the many graphs, depths, and removal fractions within a seed do not supply independent training replicates. 
Percentile bootstrap intervals with ten seeds are approximate, and counts of cells passing a test do not estimate the prevalence of the effect in an unobserved population. 
The panel expansion and primary-logit choice were dated amendments after a discarded positive-part run but before the signed crossing was read; they are therefore not an uncontaminated preregistration. The GIN margin tolerance was raised to $2\times10^{-5}$ after the Reddit-Binary--GIN numerical gap had been inspected.
The matched-superiority conjunction is exploratory and carries no across-cell error guarantee.

\paragraph{Attribution and perturbation unit.}
Integrated Gradients completeness checks the sum of attributions against the endpoints of each intervention's own path.
It does not certify individual edge scores, their ordering, or agreement with ground-truth explanations.
Our primary ranking orders directed arcs by signed attribution, and removing one arc from an undirected graph can leave the reverse channel intact. 
Grouping reciprocal arcs into bonds avoids this half-edge intervention but changes the ranked and perturbed unit, as the observed arc--bond differences confirm. 
Random tie-breaks leave the reported cell verdicts unchanged on the tested grid, but this does not make the result independent of attribution sign, output functional, target choice, or other intervention definitions. 
The predicted class on the intact graph is the fixed target; conclusions for that target need not transfer to the true class.

\paragraph{External explainer check.}
A separate pilot did not certify ranking convergence at any tested budget through 1,600 epochs and found substantial disagreement among restart prefixes. The paired evaluator comparison remains interpretable for the measured rankings because each restart's ranking is used under both evaluators. The SubgraphX comparison in Appendix~\ref{app:tables} matches selected arc count under structural deletion, while SubgraphX returns node-induced subgraphs. These exploratory comparisons do not rank the explainers overall.

\section{Conclusion}
Deletion fidelity is indexed by the intervention that removes an edge and by the output functional that reads its effect.
Proposition~\ref{prop:equivalence} states when the intervention index can be dropped, since sum aggregation with edge-set-free coefficients, as in GIN, makes the two interventions coincide.
When it cannot be dropped, crossing the two interventions measures what the choice costs.
For edge Integrated Gradients on GCN and GATv2, each ranking has the larger deletion area under its own intervention in nine of 11 cells.
For GNNExplainer, a fixed learned ranking scores differently under the two interventions in seven of 11 cells, but matched superiority of the two learned masks holds in only four.
Reporting fidelity with its intervention, output functional, target and perturbation unit, and crossing the two interventions when the question does not fix one, makes these dependencies visible instead of leaving them to an implementation default.

\subsubsection*{AI use statement}
We used large language model assistants to draft and edit parts of the manuscript and to assist with code, figure and table scripts. The authors reviewed the resulting work and take responsibility for the paper's claims.

\subsubsection*{Reproducibility statement}
Section~\ref{sec:design} and Appendix~\ref{app:protocol} specify the data split, model training, attribution, perturbation and seed-cluster analysis; Appendix~\ref{app:theory} gives the proofs. The two intervention implementations are checked end to end on GIN, where their outputs should agree. The reported crossing is reconstructed from per-cell result records, and one primary cell was recomputed from trained weights on a second machine to four significant digits. Near-tied arc scores can change order across machines, so individual deletion areas require numerical rather than bitwise tolerance.

\bibliography{references}
\bibliographystyle{iclr2027_conference}

\appendix

\section{When the two interventions coincide: proofs}
\label{app:theory}

We keep $U\subseteq E$ for the removed arcs throughout. Write $H^{(l)}_{S}$ and $H^{(l)}_{M}$ for the hidden states produced under $T_S(G,U)$ and $T_M(G,U)$, respectively, and $U_v=\{(u,v)\in U\}$ for the removed arcs entering $v$.

\paragraph{Proof of Proposition~\ref{prop:equivalence}.}
By induction on the layer index, with hypothesis $H^{(l)}_{S}=H^{(l)}_{M}$.
At $l=0$ neither intervention touches node features, so both equal $X$.
Assume $H^{(l)}_{S}=H^{(l)}_{M}=:H^{(l)}$ and fix a node $v$.
Since the hidden states agree, the messages agree arc by arc; write $\psi_{uv}=\psi^{(l)}(h^{(l)}_u,h^{(l)}_v)$.
Under $T_S$, the aggregate runs over the surviving arcs with coefficients recomputed on $E\setminus U$, and edge-set freeness gives
\[
\mathfrak{A}_{(u,v)\in E\setminus U}\big[c^{(l)}_{uv}(E\setminus U,H^{(l)})\,\psi_{uv}\big]
=\mathfrak{A}_{(u,v)\in E\setminus U}\big[c^{(l)}_{uv}(E,H^{(l)})\,\psi_{uv}\big].
\]
Under $T_M$, the aggregate runs over all arcs, with messages on $U_v$ set to zero; zero absorption gives
\[
\mathfrak{A}\Big(\{c^{(l)}_{uv}(E,H^{(l)})\,\psi_{uv}\}_{(u,v)\in E\setminus U}\uplus\{0\}^{|U_v|}\Big)
=\mathfrak{A}_{(u,v)\in E\setminus U}\big[c^{(l)}_{uv}(E,H^{(l)})\,\psi_{uv}\big].
\]
The two aggregates are equal, so $\phi^{(l)}$ returns the same state and $H^{(l+1)}_{S}=H^{(l+1)}_{M}$.
After the last layer the readouts agree, because neither intervention removes nodes. \hfill$\square$

\paragraph{The conditions are not necessary.}
The two failures can compensate each other.
Let $\mathfrak{A}$ be the mean over the $n$ incoming arcs of $v$ and let every coefficient equal the current in-degree, $c_{uv}(E)=n$.
If $s=|U_v|<n$ arcs are removed, $T_S$ gives $\frac{1}{n-s}\sum_{E\setminus U}(n-s)\psi_{uv}=\sum_{E\setminus U}\psi_{uv}$ and $T_M$ gives $\frac{1}{n}\sum_{E\setminus U}n\,\psi_{uv}$, the same value. If all $n$ arcs are removed, both aggregates are zero under the convention that an empty mean is zero. Yet the mean does not absorb zero in general and the coefficients are not edge-set free.
Proposition~\ref{prop:equivalence} is therefore a sufficient condition only.
No architecture in our panel is of this kind.

\paragraph{Proof of Proposition~\ref{prop:divergence}.}
At the first layer both runs share $H^{(0)}=X$, hence the same messages $\psi^{(0)}(x_u,x_v)$. Both sums reduce to the surviving arcs: under $T_S$ because $U$ is absent, and under $T_M$ because arcs in $U$ contribute zero.
Subtracting term by term gives Equation~\ref{eq:divergence}. \hfill$\square$

\paragraph{The bound and its limits.}
Let $R_v(U)=\sum_{(u,v)\in E\setminus U}|c^{(0)}_{uv}(E\setminus U,X)-c^{(0)}_{uv}(E,X)|$ be the total first-layer coefficient shift on arcs entering $v$. The triangle inequality applied to Equation~\ref{eq:divergence} gives $\|\Delta_v\|_\infty\le R_v(U)\max_{(u,v)\in E\setminus U}\|\psi^{(0)}(x_u,x_v)\|_\infty$, with an empty maximum taken as zero. Thus $R_v(U)=0$ for every $v$ implies $\Delta\equiv0$.
The bound does not order the divergence by the shift.
In one hidden dimension, shifts of $2$ on two messages $1$ and $-1$ give $R_v=4$ and $\Delta_v=0$, while a shift of $1$ on a single message $1$ gives $R_v=1$ and $\Delta_v=1$.
A linear readout can also project any nonzero pooled difference to zero.
A larger shift therefore need not produce a larger fidelity gap in a trained model.

\paragraph{Witnesses for the two failure modes.}
Table~\ref{tab:witness} reports the worst absolute logit gap between $T_S$ and $T_M$ over every removal set of a small irregular graph, on untrained models and with the same intervention implementations used in the experiments.
A random initialization shows that a violation \emph{can} produce divergence; it does not show that every violation must.
The last two rows share weights, graph and removal sets and differ only in the aggregator, which isolates zero absorption from coefficient dependence.

\begin{table}[h]
\centering
\small
\caption{Worst absolute logit gap between the two interventions on untrained models, over all removal sets of a small irregular graph.}
\label{tab:witness}
\begin{tabular}{lccr}
\toprule
Convolution & Absorbs zero & Edge-set free & Worst $|$gap$|$ \\
\midrule
GIN, sum, unit coefficients & yes & yes & $0.000$ \\
GraphSAGE, sum & yes & yes & $0.000$ \\
GCN, degree normalization & yes & no & $0.605$ \\
GATv2, attention & yes & no & $0.715$ \\
GraphSAGE, mean & no & yes & $0.410$ \\
\bottomrule
\end{tabular}
\end{table}

\paragraph{Scope of Equation~\ref{eq:layer} in our panel.}
GATv2 averages four attention heads; the head average divides by a fixed number of heads and does not break zero absorption.
Residual connections sit inside $\phi^{(l)}$.
Every panel cell pools by sum over nodes, and neither intervention removes nodes, so the readout depends on the final node states only.

\section{Experimental protocol}
\label{app:protocol}
\paragraph{Datasets.}
All nine datasets are graph-classification tasks.
The four synthetic benchmarks have constant node features and an edge-level ground truth.
\emph{BA-2Motifs}~\cite{luo2020parameterized} attaches a five-node house or a five-node cycle to a Barab\'asi--Albert base graph, and the label is the motif type.
\emph{BA-MultiShapes}~\cite{bamultishapes} plants combinations of a house, a grid and a wheel on a Barab\'asi--Albert base, and a graph is positive when exactly two of the three are present.
\emph{SPMotif}~\cite{spmotif} plants a class-defining motif on a base drawn from three families. We generate it at bias $1/3$, where the base is independent of the label.
\emph{Synth-Dist} is our proximity task.
Two identical five-node house motifs are planted in a Barab\'asi--Albert backbone at a controlled graph distance $d$, and the binary label is whether $d$ is below or above a threshold.
Both classes contain exactly two motifs, so graph size and node counts cannot leak the label.
The model must compose information along the backbone path between the motifs, so the receptive field needed to solve the task is a generator parameter.
The mean GIN test accuracy over ten seeds grows from $0.62$ at $K=2$ to $1.00$ at $K=5$.
The ground-truth mask marks the edges of both motifs and the shortest backbone path between them.
Among the real-world datasets, \emph{MUTAG}~\cite{debnath1991structure} and \emph{Mutagenicity}~\cite{kazius2005derivation} label molecules by mutagenicity, and \emph{BACE}~\cite{subramanian2016computational} by $\beta$-secretase-1 inhibition.
\emph{Tox21}~\cite{tox21challenge} is reduced to the SR-ARE assay to obtain a binary task.
\emph{Reddit-Binary}~\cite{yanardag2015deep} is a social benchmark without node features, to which we add degree features.

\paragraph{Training.}
Each dataset is split into 80\% training, 10\% validation and 10\% test graphs, stratified by label, with the training seed as the split seed.
All models have hidden width 128, dropout 0.2, ReLU activations and sum pooling, and GATv2 uses four attention heads.
They are trained with Adam (learning rate $3\times10^{-3}$, weight decay $5\times10^{-4}$, batch size 64) for at most 500 epochs, with early stopping after 50 epochs without improvement of the validation loss, and the weights with the lowest validation loss are kept.
Before any attribution, a health check excluded the three Tox21 cells, whose models are near-constant classifiers, with a balanced accuracy of about $0.55$ against a majority-class accuracy of $0.839$.

\paragraph{Graphs and attribution.}
Each trained model is explained on the first 50 test graphs with at most 4,000 arcs.
The number of IG steps starts at 100, or 400 for GATv2, and is doubled until the relative completeness error $|\sum_e IG_e-(f_c(\mathbf 1)-f_c(\mathbf 0))|/|f_c(\mathbf 1)-f_c(\mathbf 0)|$, pooled over eight validation graphs, is below $0.05$ for both constructions, with a cap of 8,000 steps.
The two constructions of a model share the resulting budget, which ranges from 100 to 6,400 steps across the reported cells.
Every model of every reported cell meets the tolerance, with a largest error of $0.049$.
On Tox21--GATv2, eight models miss it, so that cell would be uncertified even if it had passed the health check.
The exploratory GNNExplainer check uses PyTorch Geometric's optimizer and regularization settings. Each fixed-budget message-channel ranking is evaluated under both interventions, with its restart draw shared by the two arms.

\paragraph{Inference.}
The edge-IG crossing intervals and the GNNExplainer mask-crossing intervals use 9,999 percentile-bootstrap resamples of the ten seed-level values. The exploratory GNNExplainer evaluator intervals use 2,000 such resamples. In all analyses, depths remain paired within training seed.
The exact sign-flip test enumerates all $2^{10}$ sign patterns of the seed-level values, so its smallest attainable two-sided $p$-value is $2^{-9}\approx0.002$.

\section{Additional results}
\label{app:tables}

Table~\ref{tab:operator-crossing-results} lists the edge-IG crossing on the target logit; Table~\ref{tab:crossing-functionals} gives its interaction on the three output functionals. Table~\ref{tab:gnnexplainer-crossing} gives the values plotted in Figure~\ref{fig:contrasts}b, and Table~\ref{tab:gnnexplainer-evaluator} reports the separate fixed-ranking evaluator check.

\begin{table}[h]
\centering
\caption{Edge-IG crossing on the target logit, directed arcs, the 11 reportable GCN and GATv2 cells. $C_S=F_{SS}-F_{MS}$ and $C_M=F_{MM}-F_{SM}$ compare the two rankings under each evaluator, and $D=C_S+C_M$. A contrast is bold when its one-sided 95\% lower bound is above zero; $D$ is bold when $p_{\mathrm{BY}}<0.05$. $p_{\mathrm{BY}}$ is the exact two-sided seed sign-flip $p$-value, Benjamini--Yekutieli corrected within the primary ($^{\dagger}$) or the secondary family. The value $0.005$ is reached when all ten seeds agree in sign. The interval is the 95\% percentile seed-cluster bootstrap interval. A check mark flags matched superiority, i.e.\ both contrasts bold.}
\label{tab:operator-crossing-results}
\small
\begin{tabular}{llrrrcrc}
\toprule
 & & \multicolumn{2}{c}{Matched contrasts} & \multicolumn{3}{c}{Interaction} & \\
\cmidrule(lr){3-4}\cmidrule(lr){5-7}
Dataset & Backbone & $C_S$ & $C_M$ & $D$ & 95\% CI of $D$ & $p_{\mathrm{BY}}$ & Matched \\
\midrule
BA-2Motifs$^{\dagger}$ & GCN & $-0.548$ & $\mathbf{0.285}$ & $-0.263$ & $[-0.561,\,0.064]$ & $0.201$ &  \\
BA-MultiShapes & GCN & $\mathbf{1.491}$ & $\mathbf{1.632}$ & $\mathbf{3.123}$ & $[2.965,\,3.271]$ & $0.005$ & $\checkmark$ \\
SPMotif & GCN & $\mathbf{0.090}$ & $\mathbf{0.407}$ & $\mathbf{0.497}$ & $[0.271,\,0.766]$ & $0.005$ & $\checkmark$ \\
Synth-Dist & GCN & $-0.112$ & $\mathbf{1.243}$ & $\mathbf{1.131}$ & $[1.007,\,1.280]$ & $0.005$ &  \\
MUTAG & GCN & $\mathbf{0.438}$ & $\mathbf{0.300}$ & $\mathbf{0.738}$ & $[0.546,\,0.964]$ & $0.005$ & $\checkmark$ \\
Mutagenicity & GCN & $\mathbf{0.892}$ & $\mathbf{0.222}$ & $\mathbf{1.114}$ & $[0.964,\,1.265]$ & $0.005$ & $\checkmark$ \\
BACE$^{\dagger}$ & GCN & $\mathbf{0.484}$ & $\mathbf{0.353}$ & $\mathbf{0.838}$ & $[0.660,\,1.034]$ & $0.005$ & $\checkmark$ \\
Reddit-Binary & GCN & $\mathbf{0.572}$ & $\mathbf{0.681}$ & $\mathbf{1.253}$ & $[1.065,\,1.437]$ & $0.005$ & $\checkmark$ \\
\addlinespace
MUTAG & GATv2 & $\mathbf{0.132}$ & $\mathbf{0.165}$ & $\mathbf{0.297}$ & $[0.234,\,0.355]$ & $0.005$ & $\checkmark$ \\
Mutagenicity & GATv2 & $\mathbf{0.168}$ & $\mathbf{0.107}$ & $\mathbf{0.275}$ & $[0.221,\,0.321]$ & $0.005$ & $\checkmark$ \\
BACE$^{\dagger}$ & GATv2 & $\mathbf{0.333}$ & $\mathbf{0.200}$ & $\mathbf{0.533}$ & $[0.486,\,0.583]$ & $0.005$ & $\checkmark$ \\
\bottomrule
\end{tabular}
\end{table}

\begin{table}[h]
\centering
\caption{The edge-IG interaction $D$ on the three output functionals, directed arcs. $D$ is bold when $p_{\mathrm{BY}}<0.05$ (families as in Table~\ref{tab:operator-crossing-results}); M marks matched superiority. The last row counts bold entries and check marks. The functionals have different scales, so values compare only within a column.}
\label{tab:crossing-functionals}
\small
\begin{tabular}{llrcrcrc}
\toprule
 & & \multicolumn{2}{c}{Logit} & \multicolumn{2}{c}{Margin} & \multicolumn{2}{c}{Probability} \\
\cmidrule(lr){3-4}\cmidrule(lr){5-6}\cmidrule(lr){7-8}
Dataset & Backbone & $D$ & M & $D$ & M & $D$ & M \\
\midrule
BA-2Motifs$^{\dagger}$ & GCN & $-0.263$ &  & $-0.542$ &  & $\mathbf{0.0426}$ & $\checkmark$ \\
BA-MultiShapes & GCN & $\mathbf{3.123}$ & $\checkmark$ & $\mathbf{6.241}$ & $\checkmark$ & $\mathbf{0.2158}$ & $\checkmark$ \\
SPMotif & GCN & $\mathbf{0.497}$ & $\checkmark$ & $0.015$ &  & $0.0233$ &  \\
Synth-Dist & GCN & $\mathbf{1.131}$ &  & $\mathbf{2.303}$ &  & $\mathbf{0.1026}$ &  \\
MUTAG & GCN & $\mathbf{0.738}$ & $\checkmark$ & $\mathbf{1.354}$ & $\checkmark$ & $\mathbf{0.0969}$ & $\checkmark$ \\
Mutagenicity & GCN & $\mathbf{1.114}$ & $\checkmark$ & $\mathbf{2.225}$ & $\checkmark$ & $\mathbf{0.2459}$ & $\checkmark$ \\
BACE$^{\dagger}$ & GCN & $\mathbf{0.838}$ & $\checkmark$ & $\mathbf{1.662}$ & $\checkmark$ & $\mathbf{0.1414}$ & $\checkmark$ \\
Reddit-Binary & GCN & $\mathbf{1.253}$ & $\checkmark$ & $\mathbf{2.507}$ & $\checkmark$ & $\mathbf{0.1312}$ & $\checkmark$ \\
\addlinespace
MUTAG & GATv2 & $\mathbf{0.297}$ & $\checkmark$ & $\mathbf{0.595}$ & $\checkmark$ & $\mathbf{0.0564}$ & $\checkmark$ \\
Mutagenicity & GATv2 & $\mathbf{0.275}$ & $\checkmark$ & $\mathbf{0.549}$ & $\checkmark$ & $\mathbf{0.0841}$ & $\checkmark$ \\
BACE$^{\dagger}$ & GATv2 & $\mathbf{0.533}$ & $\checkmark$ & $\mathbf{1.062}$ & $\checkmark$ & $\mathbf{0.1514}$ & $\checkmark$ \\
\midrule
 & \textit{cells} & 10 & 9 & 9 & 8 & 10 & 9 \\
\bottomrule
\end{tabular}
\end{table}

\begin{table}[h]
\centering
\caption{GNNExplainer crossing on the target logit: structural mask versus the message-channel mask of PyG, cells with complete runs. Intervals are 90\% percentile seed-cluster bootstrap intervals recomputed from the depth-balanced seed means; a contrast is bold when the lower end is above zero. No multiplicity adjustment is applied.}
\label{tab:gnnexplainer-crossing}
\small
\setlength{\tabcolsep}{3pt}
\begin{tabular}{llrcrcrc}
\toprule
Dataset & Backbone & $C_S$ & 90\% CI & $C_M$ & 90\% CI & $D$ & Matched \\
\midrule
BA-2Motifs & GCN & $\mathbf{1.380}$ & $[1.156,\,1.625]$ & $-0.319$ & $[-0.487,\,-0.164]$ & $1.061$ &  \\
BA-MultiShapes & GCN & $-0.141$ & $[-0.207,\,-0.081]$ & $-0.705$ & $[-0.784,\,-0.625]$ & $-0.845$ &  \\
SPMotif & GCN & $\mathbf{0.083}$ & $[0.032,\,0.135]$ & $-0.259$ & $[-0.359,\,-0.154]$ & $-0.177$ &  \\
Synth-Dist & GCN & $-0.021$ & $[-0.052,\,0.010]$ & $\mathbf{0.063}$ & $[0.013,\,0.113]$ & $0.042$ &  \\
MUTAG & GCN & $\mathbf{0.046}$ & $[0.005,\,0.091]$ & $\mathbf{0.039}$ & $[0.019,\,0.066]$ & $0.085$ & $\checkmark$ \\
Mutagenicity & GCN & $\mathbf{0.380}$ & $[0.336,\,0.422]$ & $\mathbf{0.031}$ & $[0.011,\,0.050]$ & $0.411$ & $\checkmark$ \\
BACE & GCN & $\mathbf{0.230}$ & $[0.204,\,0.260]$ & $-0.021$ & $[-0.040,\,-0.002]$ & $0.209$ &  \\
Reddit-Binary & GCN & $\mathbf{0.093}$ & $[0.063,\,0.124]$ & $-0.045$ & $[-0.071,\,-0.017]$ & $0.048$ &  \\
MUTAG & GATv2 & $\mathbf{0.037}$ & $[0.029,\,0.045]$ & $0.000$ & $[-0.010,\,0.012]$ & $0.037$ &  \\
Mutagenicity & GATv2 & $\mathbf{0.105}$ & $[0.090,\,0.121]$ & $\mathbf{0.030}$ & $[0.019,\,0.040]$ & $0.135$ & $\checkmark$ \\
BACE & GATv2 & $\mathbf{0.189}$ & $[0.167,\,0.212]$ & $\mathbf{0.036}$ & $[0.023,\,0.049]$ & $0.225$ & $\checkmark$ \\
\bottomrule
\end{tabular}
\end{table}

\paragraph{Evaluator sensitivity of the PyG-GNNExplainer ranking.}
We next hold each learned PyG-GNNExplainer ranking fixed and evaluate it under the two interventions. 
This is a one-ranking-by-two-evaluator comparison, distinct from the IG construction-by-evaluation crossing.
At the fixed budget of 100 optimization epochs and five restarts per graph, all 11 reportable GCN and GATv2 cells have complete summaries (40 trained models per cell).
Table~\ref{tab:gnnexplainer-evaluator} reports the paired message-channel-minus-structural deletion-area gap for the same ranking.
The 95\% seed-cluster bootstrap interval excludes zero in seven of 11 cells on the logit (five positive, two negative), eight on the margin (six positive, two negative), and nine on the probability (six positive, three negative).
These counts are descriptive: this exploratory analysis has no multiplicity-adjusted cell tests.

\begin{table}[h]
\centering
\caption{One PyG-GNNExplainer ranking under the two evaluators. Entries are paired differences $F(r_{\mathrm{PyG}},T_M)-F(r_{\mathrm{PyG}},T_S)$, so a positive gap means a larger deletion area under message-channel suppression. Bold gaps have a 95\% percentile seed-cluster bootstrap interval that excludes zero; no multiplicity adjustment is applied.}
\label{tab:gnnexplainer-evaluator}
\small
\begin{tabular}{llrcrc}
\toprule
 & & \multicolumn{2}{c}{Logit} & \multicolumn{2}{c}{Probability} \\
\cmidrule(lr){3-4}\cmidrule(lr){5-6}
Dataset & Backbone & Gap & 95\% CI & Gap & 95\% CI \\
\midrule
BA-2Motifs & GCN & $\mathbf{0.363}$ & $[0.109,\,0.629]$ & $\mathbf{-0.0285}$ & $[-0.0475,\,-0.0098]$ \\
BA-MultiShapes & GCN & $\mathbf{-0.457}$ & $[-0.740,\,-0.170]$ & $\mathbf{-0.0366}$ & $[-0.0529,\,-0.0219]$ \\
SPMotif & GCN & $\mathbf{0.497}$ & $[0.069,\,0.929]$ & $\mathbf{0.0083}$ & $[0.0004,\,0.0159]$ \\
Synth-Dist & GCN & $0.156$ & $[-0.126,\,0.423]$ & $\mathbf{0.0223}$ & $[0.0020,\,0.0439]$ \\
MUTAG & GCN & $0.034$ & $[-0.140,\,0.183]$ & $0.0076$ & $[-0.0085,\,0.0275]$ \\
Mutagenicity & GCN & $\mathbf{0.209}$ & $[0.157,\,0.258]$ & $\mathbf{0.0570}$ & $[0.0460,\,0.0680]$ \\
BACE & GCN & $\mathbf{0.123}$ & $[0.036,\,0.231]$ & $\mathbf{0.0383}$ & $[0.0250,\,0.0518]$ \\
Reddit-Binary & GCN & $\mathbf{-0.598}$ & $[-0.718,\,-0.479]$ & $\mathbf{-0.0135}$ & $[-0.0253,\,-0.0031]$ \\
\addlinespace
MUTAG & GATv2 & $-0.007$ & $[-0.058,\,0.049]$ & $-0.0005$ & $[-0.0035,\,0.0030]$ \\
Mutagenicity & GATv2 & $\mathbf{0.019}$ & $[0.003,\,0.037]$ & $\mathbf{0.0095}$ & $[0.0041,\,0.0149]$ \\
BACE & GATv2 & $0.029$ & $[-0.001,\,0.058]$ & $\mathbf{0.0146}$ & $[0.0048,\,0.0241]$ \\
\bottomrule
\end{tabular}
\end{table}

The evaluator effect is therefore present beyond the controlled IG construction but neither its sign nor its detection is uniform across cells or output functionals. 
For example, BA-2Motifs--GCN has a positive logit gap and a negative probability gap, whereas BA-MultiShapes--GCN favors structural deletion on both functionals. 
These one-ranking data support evaluator sensitivity for the measured fixed-budget rankings. They do not themselves identify a construction-by-evaluation interaction; the separate mask crossing is displayed descriptively in Figure~\ref{fig:contrasts}b.

\paragraph{Matched-size comparison with SubgraphX.}
At the exact number of arcs selected by SubgraphX on each graph, we also compare GNNExplainer and SubgraphX under structural deletion.
This comparison covers nine of the 11 cells, because SubgraphX was not run to completion on Reddit-Binary--GCN and Mutagenicity--GATv2.
The seed-bootstrap interval for the difference in prediction-flip rate after removal favors GNNExplainer in seven of the nine completed cells; the intervals for BA-MultiShapes--GCN and Synth-Dist--GCN include zero. 
For Mutagenicity--GCN, the paired difference is $7.24$ percentage points (95\% CI $[5.18,9.54]$), while the predicted-class probability drop differs by $0.0345$ (95\% CI $[0.0193,0.0511]$). 
The corresponding logit and margin necessity intervals include zero, as do all sufficiency comparisons for that cell. 
SubgraphX yields a hard node-induced subgraph and is evaluated here only under its structural interpretation; matched arc count does not make the two explanations or their objectives equivalent. 
We therefore do not interpret these exploratory contrasts as an overall ranking of explainer quality.

\end{document}